\documentclass{llncs}
\usepackage{llncsdoc}
\usepackage{epsfig}
\usepackage{amsmath,times}

\newcommand\be{\vspace*{-3pt}\begin{equation}}
\newcommand\ee{\end{equation}\vspace*{-3pt}}
\newcommand\beq{\vspace*{-3pt}\begin{eqnarray}}
\newcommand\eeq{\end{eqnarray}\vspace*{-3pt}}

\def\bX{{\mathbf X}}  
   \def\bW{{\mathbf W}}

\newcommand\True{\hbox{T}}
\newcommand\ksep{$k$-separability\ }
\newcommand\s[1]{\mathcal{#1}}
\newcommand\vx{\vec{x}}

\begin{document}

% ====================================================================

\title{Separability is not the best goal for machine learning}
\author{W{\l}odzis{\l}aw Duch}
\institute{Department of Informatics, Faculty of Physics, Astronomy and Informatics, and Neurocognitive Laboratory, Center for Modern Innovative Technologies,\\
Nicolaus Copernicus University, Toru\'n, Poland,\\
Google: Wlodzislaw Duch}

\titlerunning{Duch: Separability is not the best goal for machine learning}

\maketitle

%====================================================================

\begin{abstract}
Neural networks use their hidden layers to transform input data into linearly separable data clusters, with a linear or a perceptron type output layer making the final projection on the line perpendicular to the discriminating hyperplane.
For complex data with multimodal distributions this transformation is difficult to learn.
Projection on $k\geq 2$ line segments is the simplest extension of linear separability, defining much easier goal for the learning process. Simple problems are 2-separable, but problems with inherent complex logic may be solved in a simple way by $k$-separable projections. 
The difficulty of learning non-linear data distributions is shifted to separation of line intervals, simplifying the transformation of data by hidden network layers. For classification of difficult Boolean problems, such as the parity problem, linear projection combined with \ksep is sufficient and provides a powerful new target for learning.
More complex targets may also be defined, changing the goal of learning from linear discrimination to creation of data distributions that can easily be handled by specialized models selected to analyze output distributions. This approach can replace many layers of transformation required by deep learning models. 
\end{abstract}

\noindent {\bf Keywords:}  
Neural networks, machine learning, neural architectures, separability, data complexity, deep learning, complex logic. 

%=======================================================

\section{Introduction}

Many popular classifiers, including MLPs, RBFs, SVMs, decision trees \cite{DudaHart01}, nearest neighbor and other similarity based methods
\cite{SBM01,SBMneural01}, require special approaches (architectures, kernels) or cannot handle at all complex problems, such as those exemplified by the parity problem: given a training set of binary strings $\{b_1,b_2 ... b_n\}$ determine if the number of bits equal to 1 is odd or even. In principle universal approximators, such as neural networks, are capable of handling such problems, and there is a whole literature on architectures and neural activation functions that enable the solution of parity problem. However, solutions proposed so far are manually designed to solve this particular problem, and thus will not work well for slightly different problems of similar kind.

Dealing with difficult learning problems like parity off-the-shelf algorithms (for example those collected in typical data mining packages) in the leave-one-out or crossvalidation tests for more than 3-bit problems give results at the baserate (50\%) level. Deep learning packages, such as TensorFlow, also have difficulty with higher-order Boolean data. Recurrent networks may solve some problems where there is a simple rule that generates the sequence of observations but general problems that are based on complex logic are not of this kind.
Knowing beforehand that the data represents parity problem allows for setting an appropriate MLP or LSTM architecture to solve it
\cite{Stork92,Brown93,Minor93,Setiono97,Lavretsky00,Arslanov02,Liu02,Torres02,Iyoda03,Wilamowski03}, but for complex logic in real cases (for example, dynamics of brain processes) it will be very difficult to guess how to choose an appropriate model. Learning Boolean functions similar to parity may indeed be a great test for methods that try to evolve neural architecture to solve a given problem, but so far no such systems are in sight.
The reason for this failure is rather simple: neural and other classifiers try to achieve linear separability, and non-linear separable data may require a non-trivial transformation that is very difficult to learn.
Looking at the image of the training data in the space defined by the activity of the hidden layer neurons \cite{duchvisI04,duchcoloring03} one may notice that a perfect solution is frequently found in the hidden space -- all data falls into separate clusters -- but the clusters are non-separable, therefore the perceptron output layer is unable to provide useful results.

Changing the goal of learning from linear separability to other forms of separability should make the learning process much easier. 
It would be very useful to break the notion of non-linearly separable problems into well defined classes of problems with increasing difficulty. This is done in the next section, where the simplest extension to separability, called \ksep is introduced. In the third section this notion is combined with linear projections and applied to the analysis of Boolean functions. Algorithms based on \ksep for general classification problems are outlined in section four, with the last section containing a final discussion.

%%%%%%%%%%%%%%%%%%%%%%%%%%%%%%%%%%%%%%%%%%%%%%
\section{k-separability}

Adaptive systems, such as feedforward neural networks, SVMs, similarity-based methods and other classifiers, use composition of vector mappings
\be
Y(\bX) = M^{(m)}(M^{(m-1)}...(M^{(2)}(M^{(1)}(\bX))...))
\ee
to assign a label $Y$ to the vector $\bX$. To be completely general direct dependence of mappings on inputs and previous transformations should be considered, for example $M^{(2)}(M^{(1)}(\bX),\bX)$, but for simplicity this will be omitted, considering only strictly layered mappings.
These mappings may include standardization, principal component analysis, kernel projections, general basis function expansions or perceptron transformations.
$\bX^{(i)}=M^{(i)}(\bX^{(i-1)})$ is the result of mapping after $i$ transformations steps.
For dichotomic problems considered below $Y=\bX^{(m)}=\pm 1$.

If the last transformation $Y=M^{(m)}(\bX^{(m-1)})$ is based on a squashed linear transformation,
for example a perceptron mapping
$Y=\tanh(\sum_i W_i\bX_i^{(m-1)})$, then the values of $Y$ are projections of $\bX$ on the  $[-1,+1]$ interval, and a perfect separation of classes means that for some threshold $Y_0$ all vectors from the $Y_+$ class are mapped to one side and from the $Y_-$ class to the other side of the interval. This means that the hyperplane $\bW$ defined in the $\bX^{(m-1)}$ space divides samples from the two classes, and $\bW\cdot\bX^{(m-1)}$ is simply a projection on the line $\bW$ perpendicular to this hyperplane, squashed to the $[-1,+1]$ interval by the hyperbolic tangent or similar function.
% origin?

\begin{figure} [bth]
\begin{center} \vspace*{-5mm}
\epsfxsize=165pt \epsfbox{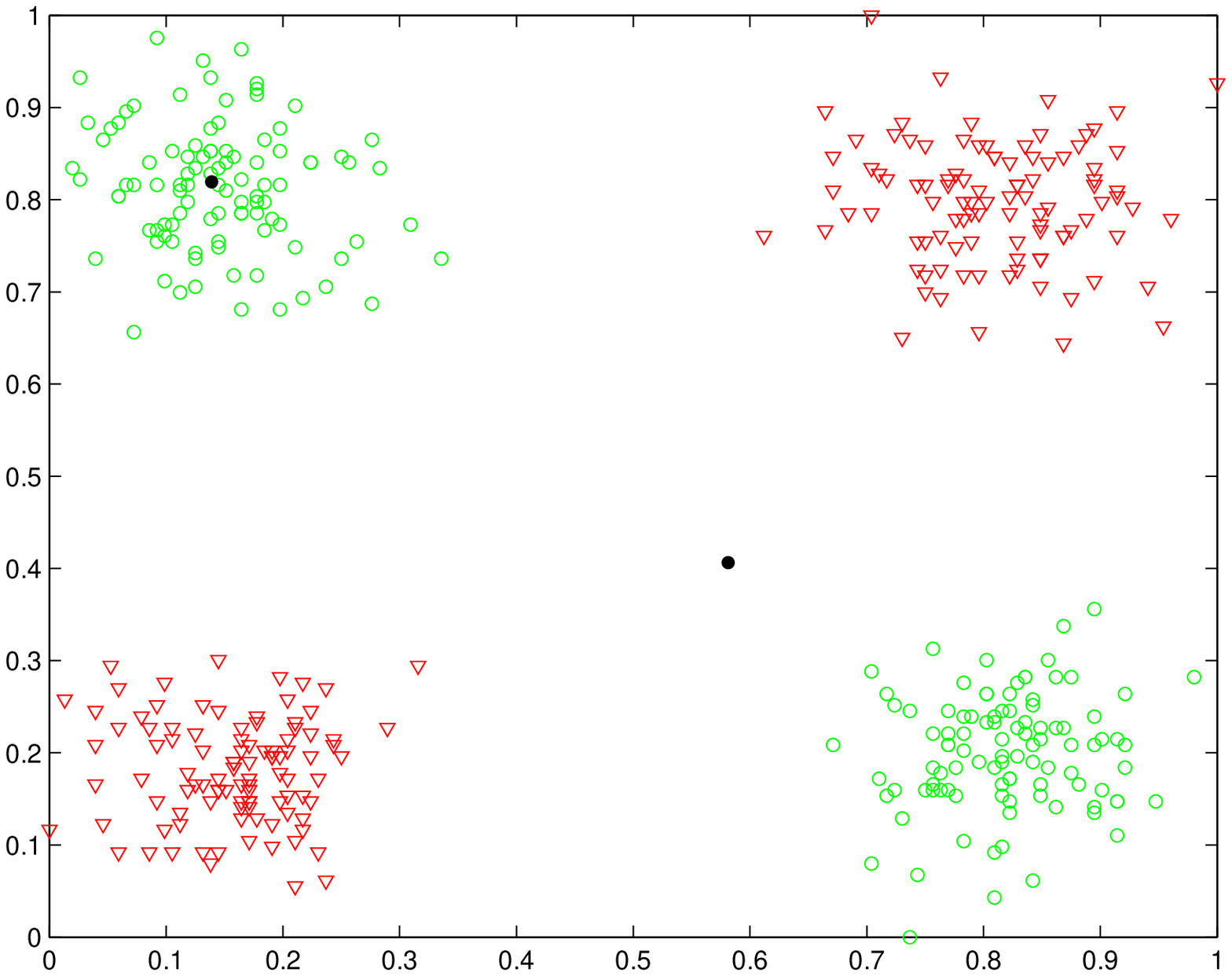}
\epsfxsize=165pt \epsfbox{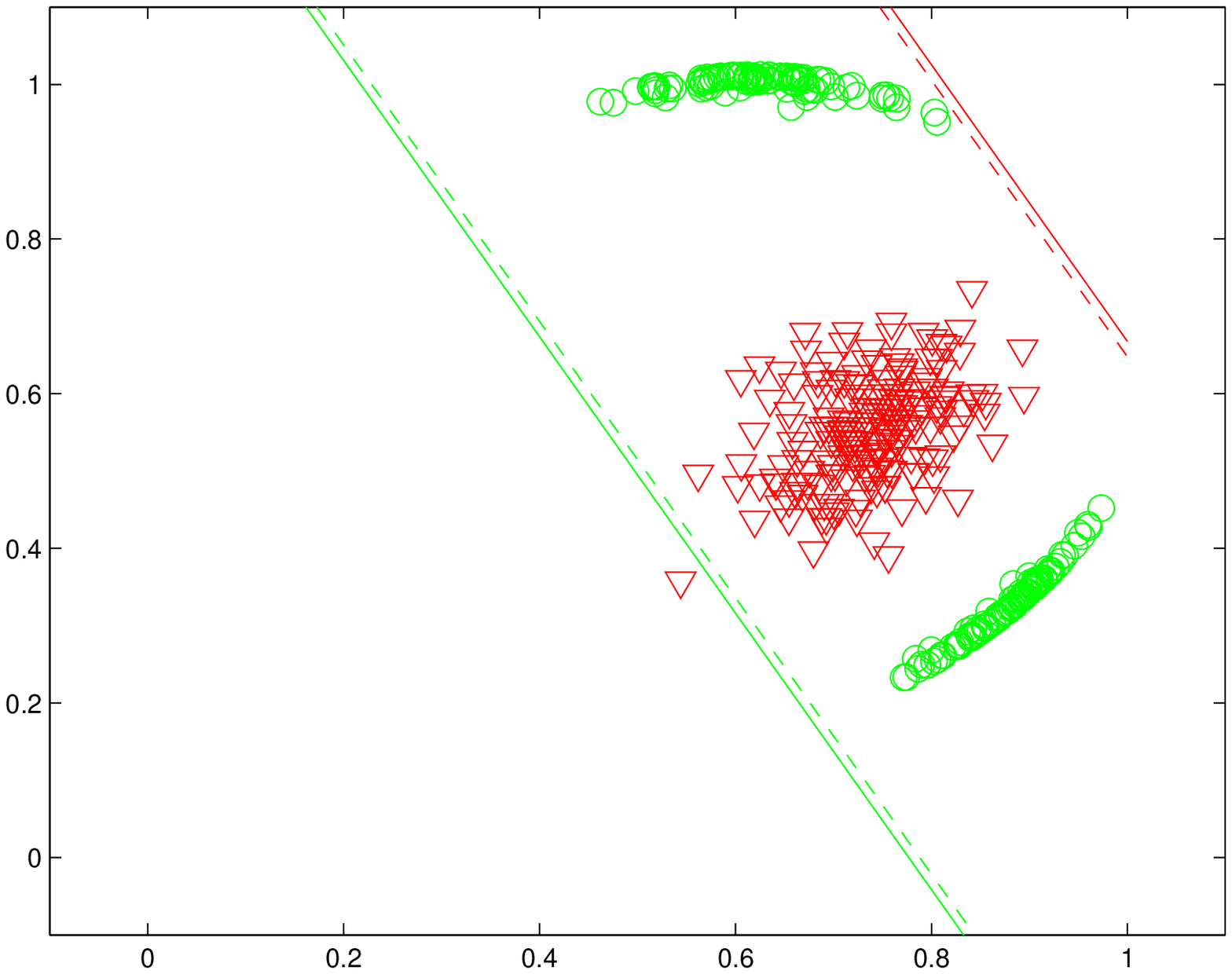}
\end{center} \vspace*{-5mm}
\caption{Noisy XOR problem solved with two Gaussian functions. Left: data distribution and position of two Gaussian functions after training using maximum likelihood principle; right: mapping of the input data to the hidden space shows perfect clusterization, showing lines representing linear weights of the two output units.}
\label{fig:rbfxor} \vspace*{-5mm}
\end{figure}

General parity problems can be solved in many ways. The simplest solution \cite{Iyoda03} is to look at the sign of the $\prod_{i=1}^n (x_i-t_i)$, with $t_i\in(0,1)$, that is to use a product neuron without any hidden neurons. This solution is very specific to the parity problem and it cannot be generalized to other Boolean functions. Many such solutions that work only for parity problem have been devised \cite{Stork92}--\cite{Wilamowski03}, but the challenge is to provide more general solutions that work also for problems of similar or higher difficulty.
%
%Parity in two dimensions is known as the XOR problem, the prototypic non-separable problem.
Many MLP training algorithms have already some difficulties to solve the XOR problem. RBF network with Gaussian hidden units cannot solve it unless special tricks are used.
%This is an example of a situation in which
Solutions based on local functions require here a large number of nodes and examples to learn, while non-local solutions may be expressed in a compact way and need only a few examples (this has been already noted in \cite{DuchJankowskiNCS99}).
Consider the noisy version of the XOR problem
%, replacing each input vector with a small Gaussian cloud
(Fig. \ref{fig:rbfxor}). RBF network with two Gaussian nodes with the same standard deviation $\sigma$ and linear output provides the following two transformations:
\beq
&& \bX\rightarrow \bX^{(1)} = (\exp(-|\bX-\mu_1|/2\sigma,\exp(-|\bX-\mu_2|/2\sigma)\\
&& \bX^{(1)}\rightarrow Y=\bX^{(2)}=\bW \cdot\bX^{(1)}
\eeq
The solution obtained using maximum likelihood approach \cite{DudaHart01} placed one basis function in the middle of left-corner cluster, and the other close to the center,
as shown in Fig. \ref{fig:rbfxor}. Although the network fails to achieve linear separability of the data, it is clear from Fig. \ref{fig:rbfxor} that all new data will be properly assigned to one of the 3 clusters formed in the hidden space; in crossvalidation test 100\% correct answers are obtained on this basis (searching for the nearest neighbor in the hidden space), while the linear output from the network achieves only 50\% accuracy (base rate). If the target $Y=\bX^{(2)}=\pm 1$ is desired the linear output provides a hyperplane (in this case a line) that tries to stay at a distance one from all data points. If a separate linear output for each class is used lines representing both outputs are parallel, with identical weights but shifted on two units, as shown in the right Fig. \ref{fig:rbfxor}. Projecting $\bX^{(1)}$ data on these lines gives one interval with the data from first class surrounded by two intervals with the data from the second class, separating the data into 3 intervals.

In $n$ dimensions a single linear unit $\bW\cdot\bX$ with all weights $W_i=1$ easily achieves separation into $n+1$ groups, with 0, 1, 2 .. $n$ bits equal to 1. This weight vector is the diagonal connecting vertices $[0,0,\dots 0]$ and $[1,1,\dots 1]$, and $\bW\cdot\bX$ is the projection on this line. Obviously using a single node with $Y=\cos(\omega \bW\cdot\bX)$ gives for $\omega=\pi$ correct answer to all parity problems, $+1$ for even and $-1$ for odd number of bits. This is the simplest general solution of the parity problem, using a single node network (this solution has not bee found previously \cite{Stork92}--\cite{Wilamowski03}). The importance of selection of appropriate transfer functions in neural networks is quite evident here (for a taxonomy of transfer functions that may be used in neural networks see \cite{transfer00} and \cite{DuchJankowskiNCS99}).
In the context of Boolean functions periodic projection is useful only for parity and its negation obtained by symmetric transformations of the hypercube with vertices labeled according to their parity. Projections of other Boolean functions may not be periodic but certainly will show several groups of vectors from alternating classes.

There is no particular reason why the target of learning should be linear separability. The last transformation $Y=M^{(m)}(\bX^{(m-1)})$ may be designed in any way that will make learning easier.
If a projection separating two clusters on a line $Y=\bW\cdot\bX$ exist the data is linearly separable. If it does not exist a projection forming 3 or more intervals containing clusters from a single class should be sought.

{\em Definition}: dataset $\{\bX_i\}$ of vectors that belong to two classes is called $k$-separable if a direction $\bW$ exist such that all points $Y_i=\bW\cdot \bX_i$ are clustered in $k$ intervals, each containing vectors from a single class.

% the data is $k$-separable if the minimum number of non-overlapping clusters that can be obtained by all possible linear projections $\bW\cdot\bX$ is equal $k$.

Linearly separable data is called 2-separable, while XOR belongs to the 3-separable category of data distributions, with projection on the $\bW=(1,1)$ line from even, odd, and again even class.
This is the simplest extension of separability, replacing the final mapping $M^{(m)}(\cdot)$ by logical rule IF $(Y\in[Y_0,Y_1]$ THEN even ELSE odd, and thus making the non-linearity rather harmless.
More sophisticated mappings from one, two or higher number of dimensions may be devised as long as transformation $M^{(m)}(\cdot)$ is easy to set up, providing easier goals for the learning process. The Error Correcting Output Codes (ECOC) \cite{ECOC95} tries also to define easier learning targets, but it is still based on linear separability, setting a number of binary targets that define a prototype ``class signature'' vectors, and comparing the distance from the actual output to these class prototypes. The change of the learning target advocated here is much more powerful.

A dataset that is \ksep may also be $(k+m)$-separable.
Strictly speaking the separability index for the data should be taken as the lowest $k$, but some learning methods may generate solutions with larger number of clusters.
For example, if the data is $k$-separable into clusters with very small number of elements, or if the margin separating the intervals between two such clusters is very small,
$(k+1)$-separability that leads to larger minimum size of small clusters and their margins may be preferred.

Solving a \ksep problem requires finding the direction $\bW$ and then setting appropriate $k-1$ thresholds defining intervals on the projection line.

{\em Conjecture}: the complexity of $k$-separable learning should be much easier then 2-separable learning.

This is rather obvious; transforming the data into \ksep form should be much easier because
additional transformations are needed to achieve linear separability, and the number of adaptive parameters may grow significantly. For example, the number of hyperplanes that an MLP network needs for the $n$-parity problem is of the order of $n$ (see comparison of solutions in \cite{Iyoda03}), giving altogether $O(n^2)$ parameters, while treating it as a $k$-separable problem requires only $n+k$ parameters. In general, cases when transformation of decision borders in the original input space $\bX$ based on continuous deformations may flatten them linear separability will be sufficient, but if discontinuous transformations are needed, as in the case of learning most Boolean functions, transformations that map data into $k$-separable form should be easier.

An interesting question is how many Boolean functions belong to the $k$-separable category. For $n$-variables there are $2^{2^n}$ possible functions; only the bounds for the number of separable Boolean functions are known: the number is between $2^{n^2-O(n)}$ and $2^{n^2}$ \cite{Zuyev89}, a vanishing fraction of all functions. Unfortunately such estimations are not yet known for the $k$-separable case.

For linearly separable data projections on $\bW$ and $-\bW$ generate symmetrical solutions $(Y_+,Y_-)$ and $(Y_-,Y_+)$; in case of $k$-separability additional symmetries and permutations are possible.

Many classification problems are quite simple and can be solved with large number of classifiers with high accuracy \cite{DuchMJ12}. $k$ value may serve as an index that can estimate complexity of the data, in contrast to estimation of flexibility of classifiers that is measured using VC-dimension or other model selection indices, such as AIC or BIC.  Consider a dataset $\s{X}= \lbrace \vx_1,\ldots,\vx_n \rbrace \subset \s{R}^d$, where each vector $\vx_i$ belongs to one of the two classes.

\begin{definition}
Dataset $\s{X}$ is called $k$-separable if a direction $\vec{w}$ exist such that all vectors projected on this direction $y_i = \vec{w}^T\vec{x}_i$ are clustered in $k$ separated intervals, each containing instances from a single class only.
\end{definition}

For example, $n$-bit parity problems are $n+1$-separable, because linear projection of binary strings exists that forms at least $n+1$ separated alternating clusters of vectors for odd and even cases. Please note that this is equivalent to a linear model with $n$ parallel hyperplanes, or a nearest-prototype model in one dimension (along the line) with $n+1$ prototypes. This may be implemented as a Learning Vector Quantization (LVQ) model \cite{Kohonen95} with strong regularization. In both cases $n$ linear parameters define direction $\vec{w}$, and $n$ parameters define thresholds placed on the $y$ line (in case of prototypes there are placed between thresholds, except for those on extreme left and extreme right, placed on the other side of the threshold in the same distance as the last prototype), so the whole model has $2n$ parameters.

It is obvious that complexity of data classification is proportional to the $k$-separability index, although for some datasets additional non-linear transformations are needed to avoid overlaps of projected clusters.
For high values of $k$ learning becomes very difficult and most classifiers, based on the Multi-Layer Perceptron (MLPs), Radial Basis Function (RBF) network, or Support Vector Machine (SVM) data models, as well as almost all other systems, including architectures used for deep learning, are not able to discover simple data models.

%%%%%%%%%%%%%%%%%%%%%%%%%%%%%%%%%%%%%%%%%%%%%%%%%%%%%%%%
\section{Boolean functions}

It is instructive to analyze in detail the case of learning Boolean functions with $n=2$ to $n=4$ bits with the simplest model based on linear projections.
Several interesting questions should be investigated:
how many \ksep cases for a given direction $\bW$ are obtained;
which direction gives the largest separation between projected clusters;
how many \ksep cases for each direction $\bW$ exist;
how many different directions are needed to find all these cases.

The Boolean functions $f(x_1,x_2,\dots x_n)\in\{-1,+1\}$ are defined on the $2^n$ vertices of $n$-dimensional hypercube. Numbering these vertices from 0 to $2^n-1$, they are easily identified converting decimal numbers to bits, for example vertex 3 corresponds to $b$-bit string 00..011.
There are $2^{2^n}$ possible Boolean functions, each corresponding to a different distribution of $\pm 1$ values on hypercube vertices. There are always two trivial cases corresponding to functions that are always true and always false, that is 1-separable functions.
Each Boolean function may be identified by a number from 0 to $2^n-1$, or a bit string from $00...0$ to $11...1$, where the value 0 stands for false or $-1$, and 1 for true or $+1$. For example, function number 9 has $2^n$ bits 00...1001, and is true only on vertex number 0 and 3.

Values of Boolean functions may be represented as black ($-1$) or white ($+1$) vertices of the hypercube. Learning a Boolean function is equivalent to separation of projections of the black and white vertices of the hypercube. Separation into small number of well separated clusters should lead to a good generalization when some function values are not known.
For two binary variables almost all non-canonical directions (not connecting vertices of the square) avoid mapping vertices of different color to exactly the same point (degeneracy) and give 6, 6 and 2 projections 2, 3 and 4-separated, respectively. It is easy to find two directions that together learn all 12 linearly separable functions (for example $\bW_{(1/3,-1/2)}$ and orthogonal direction $\bW_{(1/3,2/9)}$). These directions and $\bW_{(1,1)}$ that learns two 3-separable functions (XOR and its negation) are sufficient to learn all Boolean functions.

%%%%%%%%%%%%%%%%%%%%%%%%%%%%%%%%%%%%%%%%%%%%%%%%%%%%%%%%
\subsection{3-D case}

For 3 bits there are 8 vertices in the cube and $2^8=256$ possible Boolean functions. Functions $f(x_1,x_2,x_3)$, and their negations $\neg f(x_1,x_2,x_3)$, are related by the sign reversal symmetry or changing color of all vertices, therefore it is sufficient to consider only 128 functions corresponding to all black vertices (1 case), one black vertex (8 cases),
two blacks (${8\choose 2}=28$ cases),
three blacks (${8\choose 3}=56$ cases),
or four blacks (${8\choose 4}=70$ cases,
but only half are unique due to the sign reversal symmetry),
so together there are 1+8+28+56+35=128 such unique functions.

Projections on coordinate directions $\bW_{(001)}, \bW_{(010)}, \bW_{(100)}$ separates only three  functions $f(x_1,x_2,x_3)$$=x_k, k=1,2,3$ that belong to the 35 cases with 4 black and 4 white vertices.
There are 6 projection directions along the diagonals of the cube's faces:
$\bW_{(110)}, \bW_{(1-10)}, $$\bW_{(101)}, \bW_{(10-1)}, $$\bW_{(011)}, \bW_{(01-1)}$, and 4 projection directions along the longest diagonals of the cube:
$\bW_{(111)}, \bW_{(11-1)}, $$\bW_{(1-11)}, \bW_{(-111)}$.
Together 13 canonical directions should be considered, and a ``zero direction'' to check if there is only one class.

Consider now direction $\bW_{(110)}$.
Two points, $(0,0,0)$, $(0,0,1)$ are projected to $Y=0$,
4 points $(1,0,0), (0,1,0), (1,0,1), (0,1,1)$ are at $Y=\sqrt{2}/2$
and two points $(1,1,0)$, $(1,1,1)$ are at $Y=\sqrt{2}$.
Any Boolean function that has the same value for the first 6 points and an opposite value for the last 2 point, or vice versa, will be linearly separable. There are 4 such functions. Any function that has the same value at the middle 4 vertices, and an opposite on the remaining 4 will be 3-separable. There are 2 such functions. However, separation and size of the clusters should also be noted. For example, function 27 (00011011) is separated
by $\bW=[0.75, 1, -0.25]$ into
000 11 0 11 segments with minimum gap of 1/4
and by $\bW=[1, 0.25, -0.75]$ into
0 1 000 111 segments with the same minimum gap.
The first projection contains only one group with single 0, while the second contains two such groups, one with 0 and one with 1. For Boolean functions with small number of bits generalization is meaningless (there is no evidence to choose a particular function), but for larger number of bits avoiding small clusters should give a better chance to find most probable functions even if some values are missing. For example, for the $n$-bit parity if some of the values on vertices with $m\in[3,n-2]$ bits 1 are missing projection on $\bW=[11\dots 11]$ will still provide the best explanation of the data separating it into $n+1$ intervals.

If degeneracy is removed by slightly shifting $0, \pm1$ weight values of canonical directions
(adding 0.01 to the first, 0.02 to the second and 0.03 to the weights $\bW$ is sufficient) for an arbitrary projection direction always the same number of 1 to 8-separable functions is found:
2, 14, 42, 70, 70, 42, 14, 2.
Thus for a projection on an arbitrary direction most functions are 4 or 5-separable.
Searching for the best projection for each function using slightly perturbed canonical directions there are 2 cases of 1-separable functions, and
102, 126 and 26 of 2, 3 and 4-separable functions. For more than half of the 3-bit Boolean functions there is no linear projection that will separate the data. Almost half (126) of all functions may be learned using 3-separability. Because there are 102 linearly separable functions and each projection can recognize only 14 of them at least 8 directions are needed to check whether the function is separable.

Consider direction $\bW_{(110)}$.
Two points, $(0,0,0), (0,0,1)$ are projected to $z=0$,
4 points $(1,0,0), (0,1,0), (1,0,1), (0,1,1)$ are at $z=\sqrt{2}/2$
and two points $(1,1,0), (1,1,1)$ are at $z=\sqrt{2}$. Any boolean function that has the same value for the first 6 points and an opposite value for the last 2 point, or vice versa, will be linearly separable. There are 4 such functions. Any function that has the same value at the middle 4 vertices, and an opposite on the remaining 4 will be 3-separable. There are 2 such functions.

A summary of the number and the type of separations that each of these directions may achieve is presented in Table 1.
In general, for $n$-bit problem the number of longest diagonal hypercube directions is $2^{n-1}$, or half of the number of cube vertices (from each vertex only one such diagonal may be directed, pointing at the second vertex).

\begin{table}
\centering
\caption{Projection directions and the total number of Boolean functions that they k-separate.}
\setlength\tabcolsep{5pt}
\begin{tabular}{|c|c|c|c|c|c|}\hline
Direction & Linearly separable & 3-separable & 4-separable\\
\hline
(100)		& 2 & 0 & 0 \\	
(010)		& 2 & 0 & 0 \\	
(001)		& 2 & 0 & 0 \\	
(110)		& 4 & 2 & 0 \\	
(1-10)		& 4 & 2 & 0 \\	
(101)		& 4 & 2 & 0 \\	
(10-1)		& 4 & 2 & 0 \\	
(011)		& 4 & 2 & 0 \\	
(01-1)		& 4 & 2 & 0 \\	
(111)		& 6 & 6 & 2 \\	
(11-1)		& 6 & 6 & 2 \\	
(1-11)		& 6 & 6 & 2 \\	
(-111)		& 6 & 6 & 2 \\	
\hline
\end{tabular}\label{tab:bool3}
\end{table}

Together there is 6+4*6*2 = 54 linearly separable projections, 36 3-sep and 8 are 4-sep, so 98 functions are 4 or less separable. 
The remaining 158 functions are not separated by these projections at this level and require projection to more than one direction, or higher $k$ values.

%%%%%%%%%%%%%%%%%%%%%%%%%%%%%%%%%%%%%%%%%%%%%%%%%%%%%%%%
\subsection{4-D case}

For the 4-bit problem there are 16 hypercube vertices, with Boolean functions corresponding to 16-bit numbers, from 0 to 65535, or 64K functions. Projection on each fixed direction gives symmetric distribution of the number of \ksep functions, with the same number of functions for $k$ and $17-k$ separability. Two functions are 1-separable and two are 16-separable, changing periodically all 16 values from 0 to 1. Linear separation (and 15-separability) is found only for 30 functions,
3-separability for 210, 4 to 8 separability for 910, 2730, 6006, 10010 and 12870 functions respectively. Thus a random initialization of a single perceptron has the highest chance of creating 8 or 9 clusters in the 4-bit data.

Checking how many functions are $k$-separable requires learning the best direction for a given data. For the 4-bit case searching for the best projection along canonical directions ($W_i=0, \pm1$) that give lowest \ksep index gives 1228, 6836, 19110, 25198, 12014, 1132 and 16 projections with 2-8 clusters. These are not yet the lowest separability indices for this data, as more detailed search allowing fractional values (multiples of 1/3, 1/4, 1/5 and 1/6 in the $[-1,+1]$ range) of the $\bW$ direction coefficients shows that the highest $k$ is 5, confirming the suspicion that $k=n+1$ is the highest separability index. The number of linearly separable functions is 1880, or less than 3\% of all functions, with about 22\%, 45\% and 29\% being 3 to 5-separable. About 188 functions were found that seem to be either 4 or 5-separable, but in fact contain projection of at least two hypercube vertices with different labels on the same point. Although the percentage of linearly separated functions rapidly decreases relatively low \ksep indices resolve most of the Boolean functions.

An algorithm that searches for lowest $k$ but also maximizes minimum distance between projections of points with different labels finds projection directions (with minimum separation of 1/6 or more) that require $k=6$ for these functions and gives significantly larger separations between intervals containing vectors from a single class. With only 30 linarly separable functions per one direction and 1880 separable functions at least 63 different directions should be considered to find out if the function is really linearly separable. Learning all these functions is already a difficult problem.

For 5-bits the number of all Boolean functions grows to $2^{32}$, or over 4 billions (4G). Direct search in 5-dimensional space for each of these functions is already prohibitively expensive. It seems quite likely that for $n$-bit Boolean functions each projection direction will separate the maximum number of functions for $k\approx 2^{n}/2$, and that learning the best projection for a given function will give the largest number of functions separated into $n$ clusters, with percentage of linearly separable functions going quickly to zero. The number of elements in most cluster quickly grows, therefore with such as simple model it should be possible to learn them correctly even if only a subset of all values is given.

%%%%%%%%%%%%%%%%%%%%%%%%%%%%%%%%%%%%%%%%%%%%%%%%%%%%%%%
\section{Algorithms based on \ksep}

Linear projection combined with \ksep already gives quite powerful learning system, but
almost all computational intelligence algorithms may implement in some form \ksep as a target for learning. It is recommended to search first for linearly separable solutions, and then to increase $k$ searching for the simplest solution, selecting the best model using crossvalidation or measures taking into account the size and separation between projected clusters.
Distribution of $y(bX;\bW)$ values allows for calculation of $P(y|Y_{\pm})$ class distributions and posterior probabilities using Bayesian rules. Estimation of probability distributions in one dimension is easy and may be done using Parzen-windows kernel methods.

The main difficulty in formulating a learning procedure is the fact that targets re not fully specified; instead of a single target for $Y_+$ class two or more labels $Y_{+1},Y_{+2}$ may be needed. This may actually be of some advantage, allowing for a better interpretation of the results. It is clearly visible in the case of parity problems: each group differs not only by the parity but also by different number of 1's, providing an additional label.
Learning should therefore combine unsupervised and supervised components.
In the first step random initialization is performed several times, selecting the lowest $k$ cluster projection. Centers of these clusters $t_i, i=1\dots k$ are the target variables for learning, and each center has a class label $Y(t_i)$.
Slightly modified quadratic error function may be used for learning:
\beq
&&E(\bW,{\mathbf t}) =
{1\over 2} \sum_\bX\left(y(\bX;\bW)-t_j(\bX)\right)^2; \\ \nonumber
&& j=\arg \min_i \{ ||t_i-y(\bX;\bW)||, Y(t_i)=Y(\bX) \}
\eeq
For each input $\bX$ that belongs to the class $Y(\bX)$ the nearest (on the projected line) cluster center from the same class is taken as the learning target.
A more complex cost function may be devised that penalize for the number of clusters, for overlapping of clusters, and for impurity of clusters, but this is beyond the scope of this article.

In the two-class case there are always two possibilities: either the first class vectors are projected to the lowest $Y$ values, leading to clusters $Y_+, Y_-, Y_+, \dots$,
or vice versa, $Y_-, Y_+, Y_-, \dots$.
The 3-separable case is particularly simple and often encountered in practice. If vectors from one of the classes represent unusual objects or states (for example hypo and hiper-activity in some medical problems) projections with clusters $Y_-, Y_+, Y_-$ are fairly common. This may be checked quite easily visualizing distribution of activations for a single perceptron (linear neuron is sufficient). Additional transformations (network layers) are needed to reach linear separability, but 3-separability may often be reached using just one node.

For $k=3$ these projections are in 3 intervals: $[-\infty,a], [a,b], [b,+\infty]$.
Taking $t=(a+b)/2$, and denoting $Y_X=Y(\bX)$ a linear error function suitable for learning is:

\beq
E(a,b,\bW) &&=\sum_\bX\large[\True(y\leq t)
	{\delta(Y_X,Y_+)\max(0,y-a)+\delta(Y_X,Y_-)\max(0,a-y) }\nonumber \\
&&+\True(y>t){\delta(Y_X,Y_+)\max(0,b-y)+\delta(Y_X,Y_-)\max(0,y-b) } \large]
\eeq
where $\True(y>t)$ is 0 if false and 1 if true. This function admits a trivial $\bW=0$ solution, therefore either a condition $||\bW||$ should be introduced, or a distance scale should be fixed by requiring one of the components to be constant. It assumes that $Y_+$ vectors contribute to errors only outside of the $[a,b]$ interval, with the error growing in a linear way, and that $Y_-$ vectors contribute to error in the linear way only inside this interval. It requires good initialization to map all $Y_+$ vectors to correct side of $t$. Using this function for 3-separable Boolean functions with multiple starts to find approximate 3-separability projection quickly learns such functions using a simple gradient method. To avoid threshold functions $\True(y>t)$ may be replaced by a logistic function $\sigma(y-t)$.

\begin{figure} [t]
\begin{center} %\vspace*{-11mm}
\epsfxsize=380pt\epsfbox{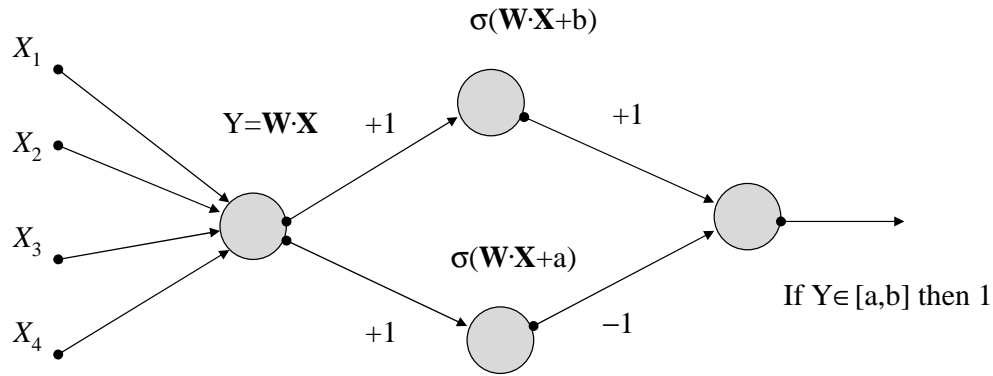}
\end{center} % \vspace*{-5mm}
\caption{MLP solution based on 3-separability assumption.}
\label{fig:MLP3sep} \vspace*{-5mm}
\end{figure}

3-separable backpropagation learning in purely neural architecture requires a single perceptron for projection plus a combination of two neurons creating a ``soft trapezoidal window'' type of function $F(Y;a,b)=\sigma(Y+a)-\sigma(Y+b)$ that implements interval $[a,b]$ \cite{duchtnn01,duchieee04}.
These additional neurons (Fig. \ref{fig:MLP3sep}) have fixed weights ($+1$ and $-1$) and biases $a, b$, adding only 2 adaptive parameters. An additional parameter determining the slope of the window shoulders may be introduced to scale the $Y$ values. The input layer may of course be replaced by a hidden layer that implements additional mapping.

This network architecture has $n+2$ parameters and is able to separate a single class bordered by vectors from other classes. For $n$-dimensional problem with 3-separable structure standard architecture requires at least two hidden neurons connected to an output neuron with $2(n+1)+3$ parameters.
For \ksep case this architecture will simply add one additional neuron for each new interval, with one bias parameter. $n$-bit parity problems require only $n$ neurons (one linear perceptron and $n-1$ neurons with adaptive biases for intervals), while in the standard approach $O(n^2)$ parameters are needed \cite{Iyoda03}. It works quite well for random Boolean problems creating small and accurate networks \cite{Groch08,Groch08a}.

%%%%%%%%%%%%%%%%%%%%%%%%%%%%%%%%%%%%%%%%%%%%%%%%%%%%%%%%%
\section{Discussion and open problems}

A radically new approach to learning has been proposed, simplifying the process by
changing the goal of learning to easier target and handling the remaining nonlinearities with well defined structure.

\ksep is a powerful concept that will be very useful for computational learning theory, breaking the space of non-separable functions into subclasses that may be separated into more than two parts. Even the simplest linear realization of \ksep with interval nonlinearities is quite powerful, allowing for efficient learning of difficult Boolean functions. So far there are no systems that can routinely handle such cases, despite a lot of effort devoted to special Boolean problems, such as the parity problem. In this paper new solutions (based on linear projections and neural architectures) have been proposed not only to the parity problem, but also to learning all Boolean functions.

This type of algorithms may have biological justification providing better explanation of the learning processes than error backpropagation or similar techniques \cite{Duch07}. Neurons in association cortex form strongly connected microcuircuits found in minicolumns, resonating with different frequencies when an incoming signal $X(t)$ appears. A perceptron neuron observing the activity of a minicolumn containing many microcircuits learns to react to signals in an interval around particular frequency.
Combination of outputs from selected perceptron neurons is used to discover a category. These outputs may come from resonators of different frequencies, implementing an analogue to the combination of disjoint projections on the $\bW\cdot\bX$ line.

Redefining the goal of learning is an interesting concept that creates many open problems.
How many boolean function each direction $k$-separates in general case?
What minimal $k$ is sufficient for $n$-bit problems?
How will different cost functions perform in practice?
What other simple ways to ``disarm'' linearites, besides projection on a $k$-segment line, may be used? 
How to use these ideas in deep learning architectures? 

Some of these questions have been addressed in papers cited below (\cite*{Duch07, Duch07a, Duch11, Groch07--Groch11}, including constructive neural networks based on the \ksep idea, and novel targets for learning that are not based on simple projections. 
Interesting application of \ksep to collaborative recommender systems has been published by Alexandridis, Siolas, and Stafylopatis \cite{Alexandridis10,Alexandridis12}

{\bf Remark:} This paper is an extension of my original paper introducing \ksep idea \cite{Duch06}, and will be extended to include new development of this concepts. 

%%%%%%%%%%%%%%%%%%%%%%%%%%%%%%%%%%%%%%%%%%%%%%%%%%%%%%%%%%%%%%%%%

\end{document}